\def\year{2021}\relax
\documentclass[letterpaper]{article} 
\usepackage{aaai21}  
\usepackage{times}  
\usepackage{helvet} 
\usepackage{courier}  
\usepackage[hyphens]{url}  
\usepackage{graphicx} 
\usepackage{natbib}  
\usepackage{caption} 
\usepackage{eufrak}
\usepackage{amsmath}
\usepackage{amssymb}
\usepackage{multirow}
\usepackage{colortbl}
\usepackage{xcolor}

\title{Frequency Consistent Adaptation for Real World Super Resolution}
\author{
Xiaozhong Ji\textsuperscript{1}\thanks{~indicates equal contribution.}, Guangpin Tao\textsuperscript{1}\footnotemark[1], Yun Cao\textsuperscript{2}, Ying Tai\textsuperscript{2}, Tong Lu\textsuperscript{1}\thanks{~Corresponding author.},\\
Chengjie Wang\textsuperscript{2}, Jilin Li\textsuperscript{2}, Feiyue Huang\textsuperscript{2}\\
}

\affiliations{
\textsuperscript{1}National Key Lab for Novel Software Technology, Nanjing University\\
\textsuperscript{2}Tencent Youtu Lab\\
shawn\_ji@163.com, tgpin@smail.nju.edu.cn, \{yuncao, yingtai\}@tencent.com,\\
lutong@nju.edu.cn, \{jasoncjwang, jerolinli, garyhuang\}@tencent.com \\

}

\begin{document}

\maketitle

\begin{abstract}
Recent deep-learning based Super-Resolution (SR) methods have achieved remarkable performance on images with known degradation. However, these methods always fail in real-world scene, since the Low-Resolution (LR) images after the ideal degradation (\textit{e.g.}, bicubic down-sampling) deviate from real source domain. The domain gap between the LR images and the real-world images can be observed clearly on frequency density, which inspires us to explictly narrow the undesired gap caused by incorrect degradation. From this point of view, we design a novel Frequency Consistent Adaptation (FCA) that ensures the frequency domain consistency when applying existing SR methods to the real scene. We estimate degradation kernels from unsupervised images and generate the corresponding LR images. To provide useful gradient information for kernel estimation, we propose Frequency Density Comparator (FDC) by distinguishing the frequency density of images on different scales. Based on the domain-consistent LR-HR pairs, we train easy-implemented Convolutional Neural Network (CNN) SR models. Extensive experiments show that the proposed FCA improves the performance of the SR model under real-world setting achieving state-of-the-art results with high fidelity and plausible perception, thus providing a novel effective framework for real-world SR application. 
\end{abstract}

\section{Introduction}

Super-Resolution (SR) is a basic low-level visual problem~\cite{freeman2002example, glasner2009super}, which is defined as enlarging the resolution of a Low-Resolution (LR) image and restoring it to a High-Resolution (HR) image. 
In recent years, deep-learning methods have dominated the research in the SR field, and lots of novel structures~\cite{dong2015image, tai2017memnet, chen2018fsrnet, tai2017image, lai2017deep, kim2016accurate, lim2017enhanced} are proposed to improve performance on standard benchmarks.
However, known degradation used to train these models is not suitable to real-world scenarios. 
In fact, the SR model is sensitive to different degradation~\cite{zhang2019deep, zhou2019kernel, gu2019blind}. 
Inconsistent degradation leads to generating undesirable SR results, either losing high-frequency detail or producing artifacts.  
\begin{figure}[t!]
\centering
\includegraphics[width=0.74\columnwidth]{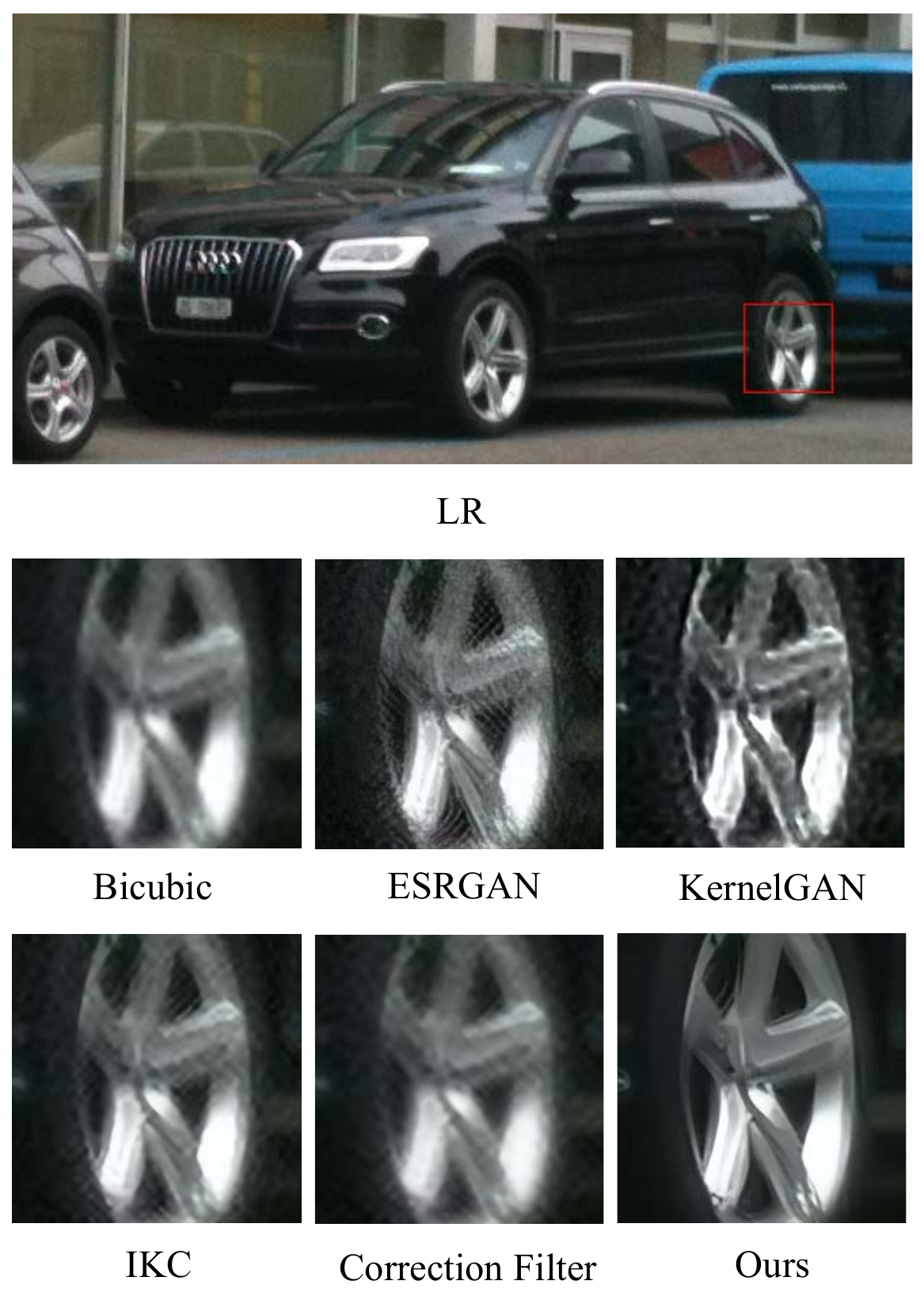} 
\caption{\textbf{$\mathbf{4\times}$ SR results of different methods on a real image taken by smartphone.} The red rectangle area of each result is cropped and enlarged for visual comparison. Compared with other results, our method generates natural HR image without blur or artifacts.} 
\label{fig:shortshow}
\end{figure}

To address the challenge of Real-World Super-Resolution (RWSR), some non-blind/blind SR methods have been proposed to improve SR performance under real degradation. Among them,  ZSSR~\cite{shocher2018zero} explores the internal information inside the image and uses the zero-shot learning method by training on the test image. However, the given kernel is needed when constructing training pairs, which is not available in real scene. 
IKC~\cite{gu2019blind} tries to construct supervised data and predict the degraded kernel from LR images. However, the supervised training is not suitable for real-world scene due to the lack of ground-truth blur kernel that can be obtained. KernelGAN~\cite{bell2019blind} utilizes the prior of recurrent patches across scales in natural LR images to train a deep linear network, whose convolution kernels can be calculated to obtain the estimated kernel. However, its insufficient recurrent patch prior and complex implementation with manual constraints limit its accuracy. 

To better deal with RWSR, we propose a novel method named Frequency Consistent Adaptation (FCA). 
By accurately estimating the degradation inside the source image, our FCA can generate realistic HR-LR image pairs. 
We design an adaptation generator to learn the blur kernel from the source image, and then degrade the down-sampled LR image. 
In order to draw the degraded LR image as close as possible to the source domain, we propose a novel Frequency Density Comparator (FDC). The frequency distributions of different degraded images show obvious differences, and their orders motivate us to construct the correspondence between the frequency domain and the blur kernel. FDC can learn embeddings related to frequency density which are essential to distinguish images with different degrees of degradation. Through trained in a self-supervised way, FDC provides effective gradient information for the adaptation generator. By constructing training images consistent to the real degradation, FCA can be combined with existing SR networks to improve their performance. Figure~\ref{fig:shortshow} shows an example of our FCA's performance improvement for real-world SR. Compared with the existing state-of-the-art methods, our result achieves higher visual quality.

In summary, our overall contribution is three-fold:
\begin{itemize}
\item We propose a novel frequency consistent adaptation for real-world super-resolution, which guarantees frequency consistency for realistic degradation.
\item We carefully design frequency density comparator to provide guidance for accurate blur kernel estimation. Our unsupervised training strategy is flexible for real scene.
\item Extensive experiments on various synthetic and real-world datasets show that the proposed FCA achieves state-of-the-art performance.
\end{itemize}

\section{Related Work}

\paragraph{CNN-based Super Resolution}
Recently, many Convolutional Neural Networks (CNN) based SR networks~\cite{lim2017enhanced, pan2018learning, he2019ode, hu2019meta, li2019feedback, qiu2019embedded,yin2019fan} achieve strong performance on bicubic down-sampled images. Among them, EDSR~\cite{lim2017enhanced} adopts a deep residual network for training SR model.
Zhang~\cite{zhang2018image} proposes a residual in residual structure to construct very deep network that achieves better performance than EDSR.
Haris~\cite{haris2018deep} proposes deep back-projection networks to exploit iterative up- and down-sampling layers, providing an error feedback mechanism for projection errors at each stage.

\begin{figure}[t!]
\centering
\includegraphics[width=0.9\columnwidth]{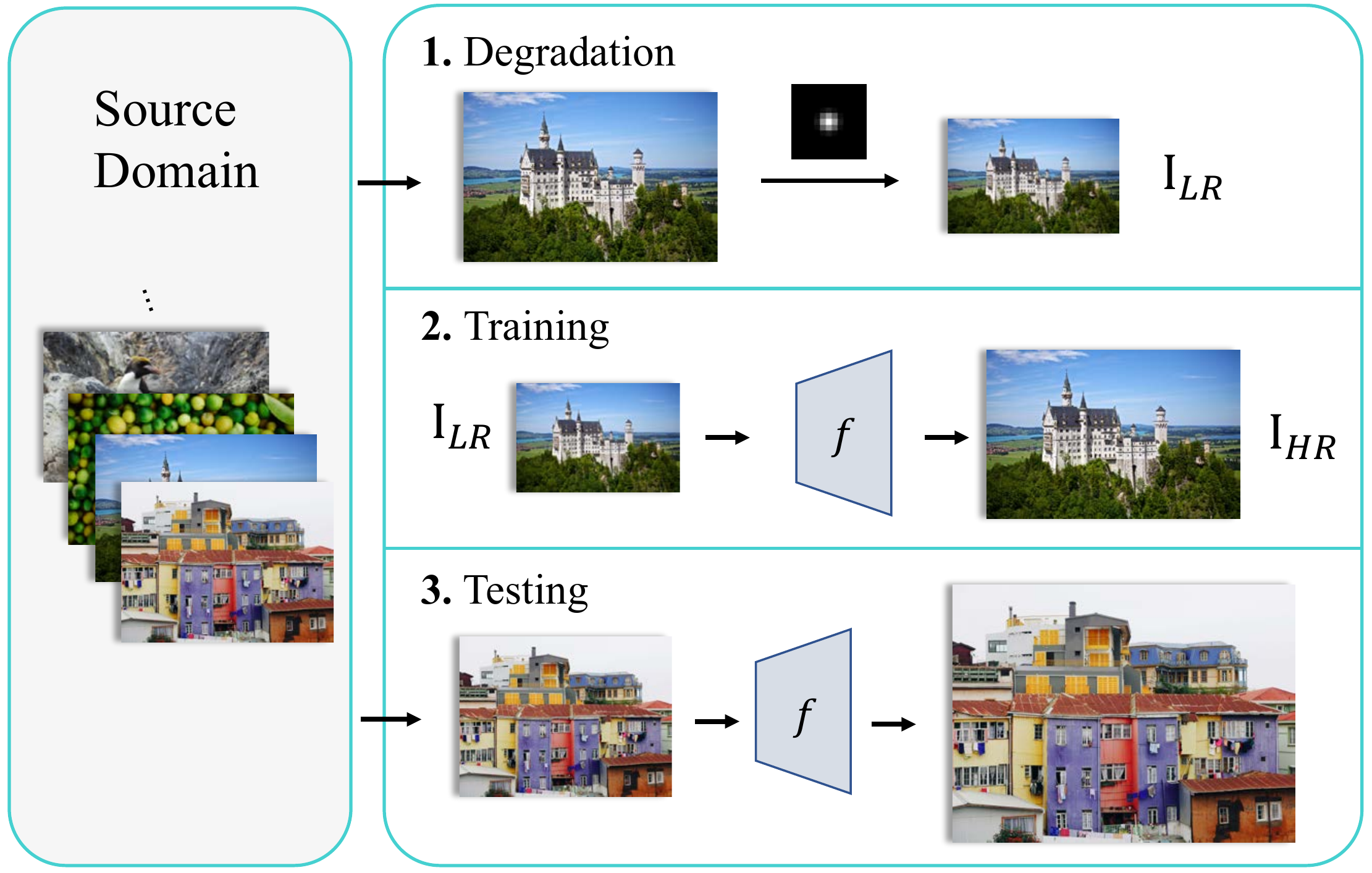} 
\caption{\textbf{Training and testing pipeline of real-world super-resolution.} When applying a SR model to real scene, a typical pipeline includes: 1. Constructing LR-HR pair by performing degradation with a blur kernel. 2. Training a SR model $f$ to reconstruct HR image ${\mathbf I}_{HR}$ for each ${\mathbf I}_{LR}$. 3. Testing SR model on real-world image.} 
\label{fig:rwsr}
\end{figure}

\begin{figure}[t!]
\centering
\includegraphics[width=0.9\columnwidth]{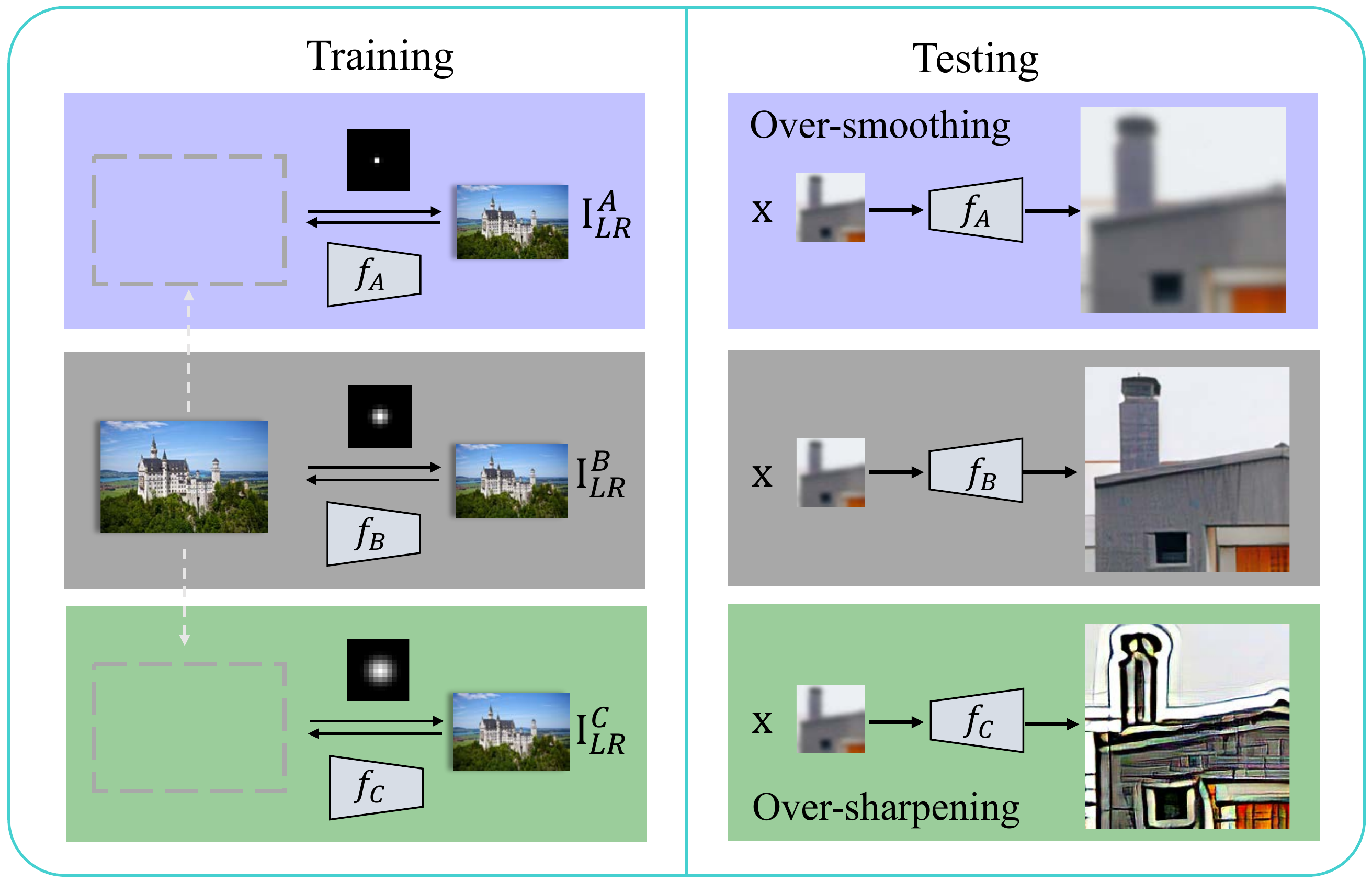} 
\caption{\textbf{Illustration of the degradation's importance in SR.} Different blur kernels might get different degraded LR images, resulting in different SR models. For a degradation-unknown image ${\mathbf x}$, only SR model ($f_B$) with a consistent degradation kernel generates natural HR result (Middle row), while those ($f_A$ and $f_C$) with inconsistent kernels generate over-smoothing (Top row) or over-sharpening (Bottom row) images.}
\label{fig:degradation}
\end{figure}

Although these works have achieved good performance with respect to fidelity, the generated images have poor visual effects and appear blurry. To address this issue, some researchers propose to enhance realistic texture via spatial feature transform~\cite{zhang2019image, zheng2018crossnet, wang2018recovering}. 
Furthermore, some Generative Adversarial Networks (GAN) based methods~\cite{ledig2017photo, zhang2019ranksrgan, wang2018esrgan} pay more attention to visual effects, introducing adversarial losses~\cite{goodfellow2014generative} and perceptual losses~\cite{johnson2016perceptual}. 
Soh~\cite{soh2019natural} proposes a natural manifold discrimination to classify HR images with blurry or noisy images, which is used to supervise the quality of the generated images.
However, these SR models trained on the data generated by bicubic kernel can only work well on ideal dataset which is \textit{inconsistent} with real-world needs. In this paper, we proposed to analyse the degradation of real images, and achieve robust performance in real scene. 

\paragraph{Real World Super Resolution} 
For real-world super-resolution application~\cite{lugmayr2020ntire, ji2020real}, SR models need to be tested on real scene which is often deviated from ideal domain.
To overcome these challenges, recent works with new training strategy have been proposed. Despite the difference in detail, the common strategy of these models for RWSR is described in Figure~\ref{fig:rwsr}.
These methods~\cite{zhou2019kernel} are trained on the artificially constructed degraded training pair, which further enhanced the robustness of the SR model. 
However, the explicit modeling way adopted by these methods needs sufficient prior about degradation, therefore the scope of application is limited. Another problem is that these methods~\cite{zhang2018learning,zhang2019deep} evaluate performance on self-made datasets, which lacks objective and fair comparison on public benchmark datasets. 
In this work, we mainly focus on kernel estimation, and the noise estimation can be easily introduced as~\cite{ji2020real}. We show that different degrees of kernel degradation have a huge impact on the performance of the SR model in Figure~\ref{fig:degradation}.

To achieve better performance on real scene images, several recent works considering degradation have been proposed.
ZSSR~\cite{shocher2018zero} abandons the training process on external data and train a specific model for each test image with more attention to the internal information of the image. However, as a non-blind model, ZSSR needs given blur kernel, which restricts its applicability. 
KernelGAN~\cite{bell2019blind} propose unsupervised kernel estimation GAN to generate down-sampling kernel. It can be used as given kernel for ZSSR. However, KernelGAN adopts only adversarial loss thus often produces unstable results.
IKC~\cite{gu2019blind} proposes to explicitly predict blur kernel in an iterative way. The supervised training method only work for synthetic data.
Our unsupervised FCA is not limited to synthetic images but also suitable for real-world images.
Correction Filter~\cite{hussein2020correction} modifies the low-resolution image to match one that is obtained by another kernel (\textit{e.g.}, bicubic) and thus improves the results of existing pre-trained CNNs. However, degraded LR often loses important frequency information, which makes it hard to rematch bicubic domain. 
In contrast, we match the LR domain to source domain without necessary to reconstruct important details. Instead, we encourage CNN models to generate rich high-frequency details in training phase.

\section{Frequency Consistent Adaptation}

\paragraph{Problem Formulation}
Generally, we assume LR image is obtained from HR image by the following degradation:
\begin{equation}
{\mathbf I}_{LR} = ({\mathbf I}_{HR} {\downarrow}_s) \otimes \mathbf {k}  + \mathbf {n},
\end{equation}
where ${\mathbf k}$ and ${\mathbf n}$ indicate blur kernel and noise, respectively. Based on the degraded LR-HR pairs, the ideal SR model for source domain should be:
\begin{equation}
f^* = \arg \min _f \{ \mathbb{E} [ \| \mathbf {I}_{HR}  - f({\mathbf I}_{LR}) \| ]\},
\end{equation}
where $f$ denotes SR model.

\paragraph{Frequency Consistency}

\begin{figure}[t!]
\centering
\includegraphics[width=0.95\columnwidth]{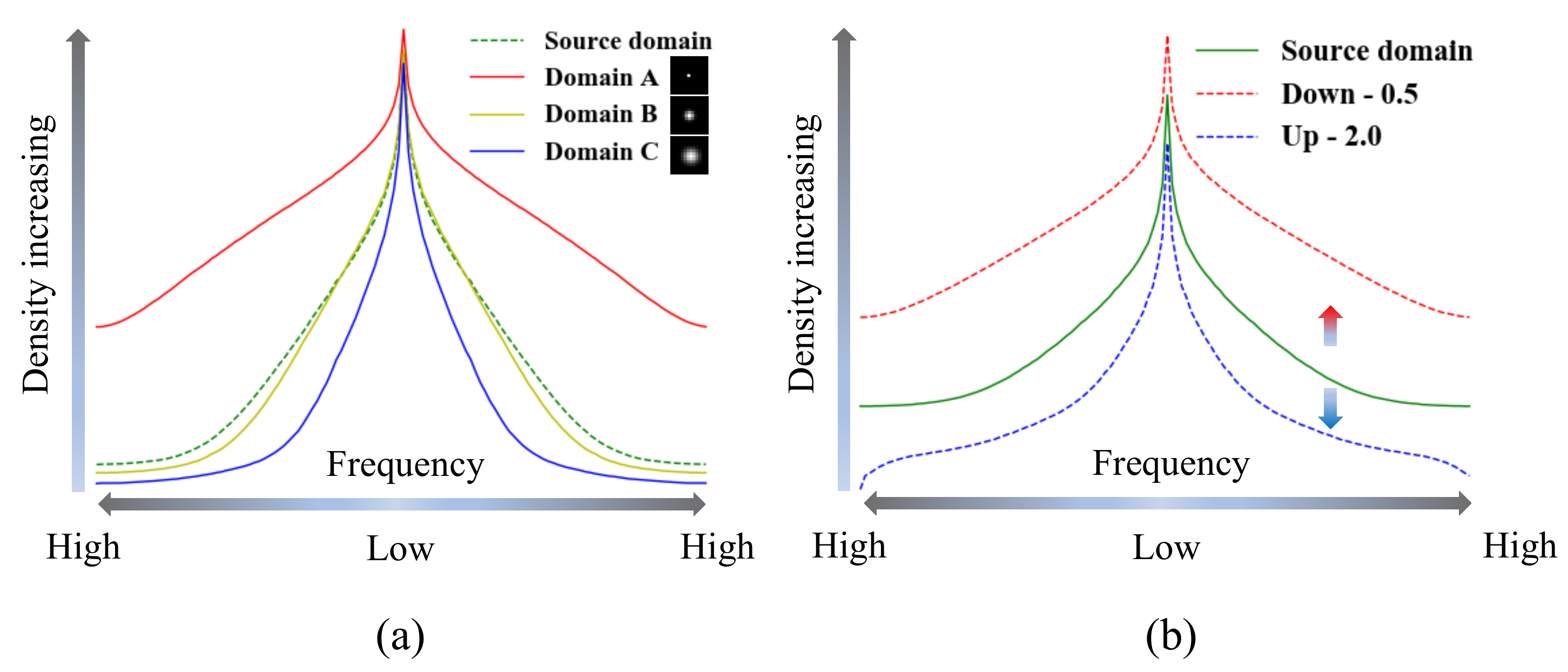} 
\caption{(a) \textbf{Degradation with different blur kernels show difference in frequency density.} The frequency is increasing from middle of horizon axis. Domain A/C represents images degraded with small/big variance blur kernel, and it has a strong/weak density in frequency domain. Domain B is a consistent degradation, thus is close to source domain. (b) \textbf{Impacts of down- and up-sampling on frequency density.} Down-sampling on source domain with scale factor $0.5$ might increase frequency density, and up-sampling with $2.0$ impacts the opposite way.}
\label{fig:freq}
\end{figure}

We observe the evidence that frequency density of LR images is related to the corresponding degradation shown in Figure~\ref{fig:freq} (a).
We calculate frequency density as 
\begin{equation}
\begin{aligned}
\mathcal{F}_X(l) &=  \frac{1}{N} \Sigma_{x\in X}{|F_l(x)|},
\end{aligned}
\end{equation}
where $\mathcal{F}_X(l)$ denotes density of frequency $l$ on domain $X$ with $N$ images. We average the two-dimensional Fourier transform along a certain dimension to get $F_l(\cdot)$. 
The relationship between degradation and frequency density motivates us to keep frequency consistency between ${\mathbf I}_{LR}$ and source image ${\mathbf x}$. 
We focus on estimating ${\mathbf k}$ with frequency domain regularization, which can be formulated as
\begin{equation}
\begin{aligned}
{\mathbf k}^* = \arg \min _{\mathbf k}   \Phi (({\mathbf I}_{HR} {\downarrow}_s) ~ \otimes {\mathbf k}, {\mathbf x}),
\end{aligned}
\end{equation}
where ${\mathbf x}$ indicates image from source domain, and $\Phi$ represents frequency regularization. 
However, it is hard to directly involve Fourier transform into networks. Guided by the proposed frequency consistency losses, FCA provides LR images that are frequency-consistent with images in source domain $S$. The HR images can be obtained directly from the source images ${\mathbf x} \in S$ or by performing down-sampling or deblurring. 
Those constructed HR-LR training pairs can then be used to train SR models specifically for $S$.

\begin{figure*}[!ht]
\centering
\includegraphics[width=0.8\textwidth]{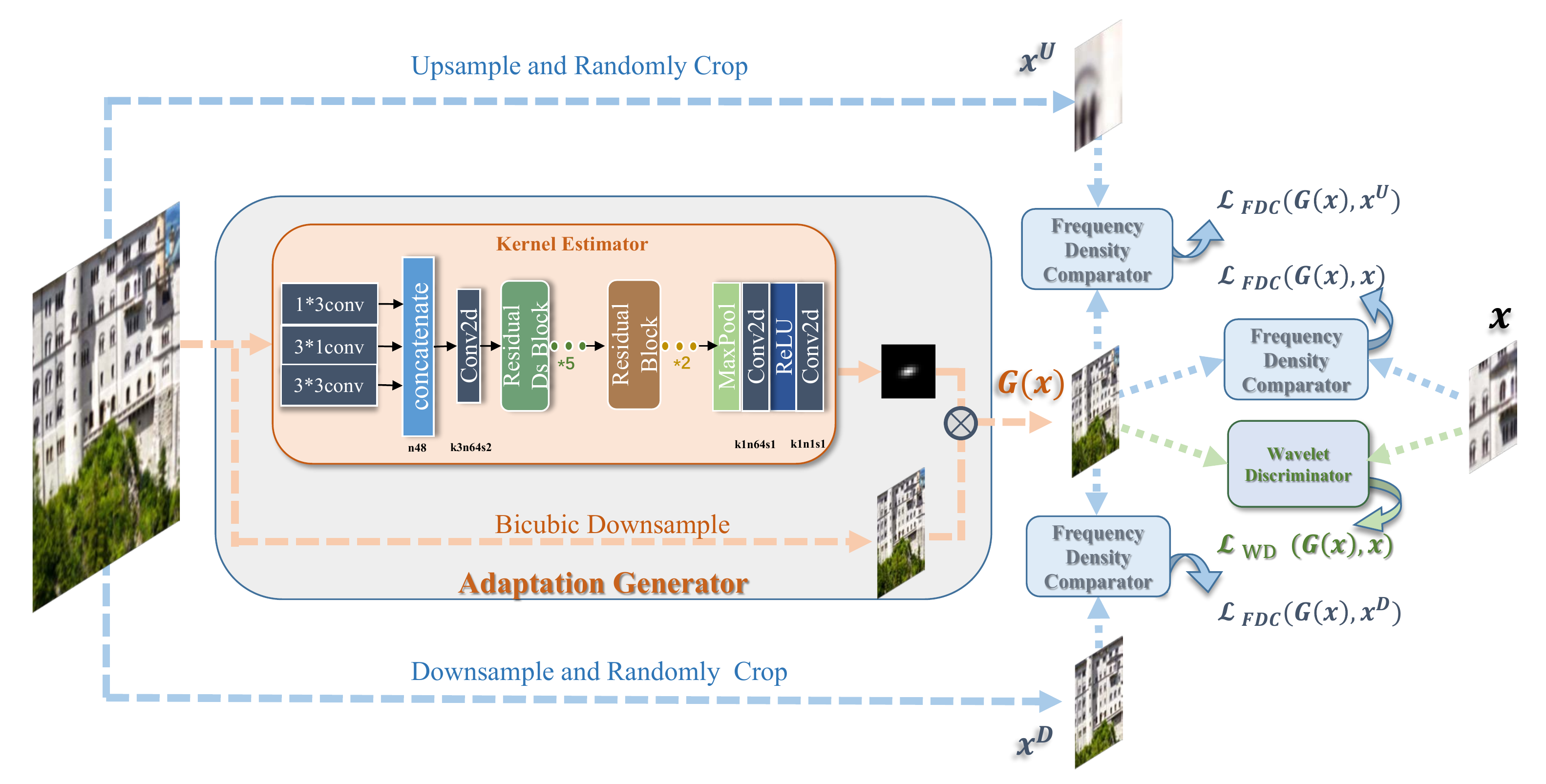} 
\caption{\textbf{Framework of the proposed frequency consistent adaptation.} The adaptation generator takes the unknown degraded image as input, and generates LR image G(x), while the frequency density comparator and the wavelet discriminator provide frequency consistency losses to guide the adaptation generator.} 
\label{fig:estimator}
\end{figure*}


\paragraph{Overall Framework}

Our FCA shown in Figure~\ref{fig:estimator} consists of three components: the adaptation generator, FDC, and the wavelet discriminator. The adaptation generator generates degraded LR with the same frequency density according to the input image, which is optimized by the guidance of FDC module and the wavelet discriminator module. 


\subsection{Adaptation Generator}
For an input source image ${\mathbf x}$, the adaptation generator first analyzes the degradation and outputs an anisotropic Gaussian kernel. Then the kernel is convolved with the down-sampled image of ${\mathbf x}$ with scale factor $s$ to generate LR image $G(x)$, which is formulated as
\begin{equation}
\begin{aligned}
G(x) = ({\mathbf x}\downarrow_s) \otimes {\mathbf k} ({\mathbf x}),
\end{aligned}
\end{equation}
where ${\mathbf k} ({\cdot})$ means the kernel estimator. More precisely, we describe the estimated kernel with three parameters:
\begin{equation}
\begin{aligned}
{\mathbf k} ({\mathbf x}) =  g(r_1({\mathbf x}), r_2({\mathbf x}), \theta({\mathbf x})),
\end{aligned}
\end{equation}
where $r_1$, $r_2$, $\theta$ indicate the horizontal, vertical radius and the angle of rotation respectively. $g(\cdot)$ denotes anisotropic Gaussian kernel.

\subsection{Frequency Density Comparator}

\begin{figure}[t!]
\centering
\includegraphics[width=0.9\columnwidth]{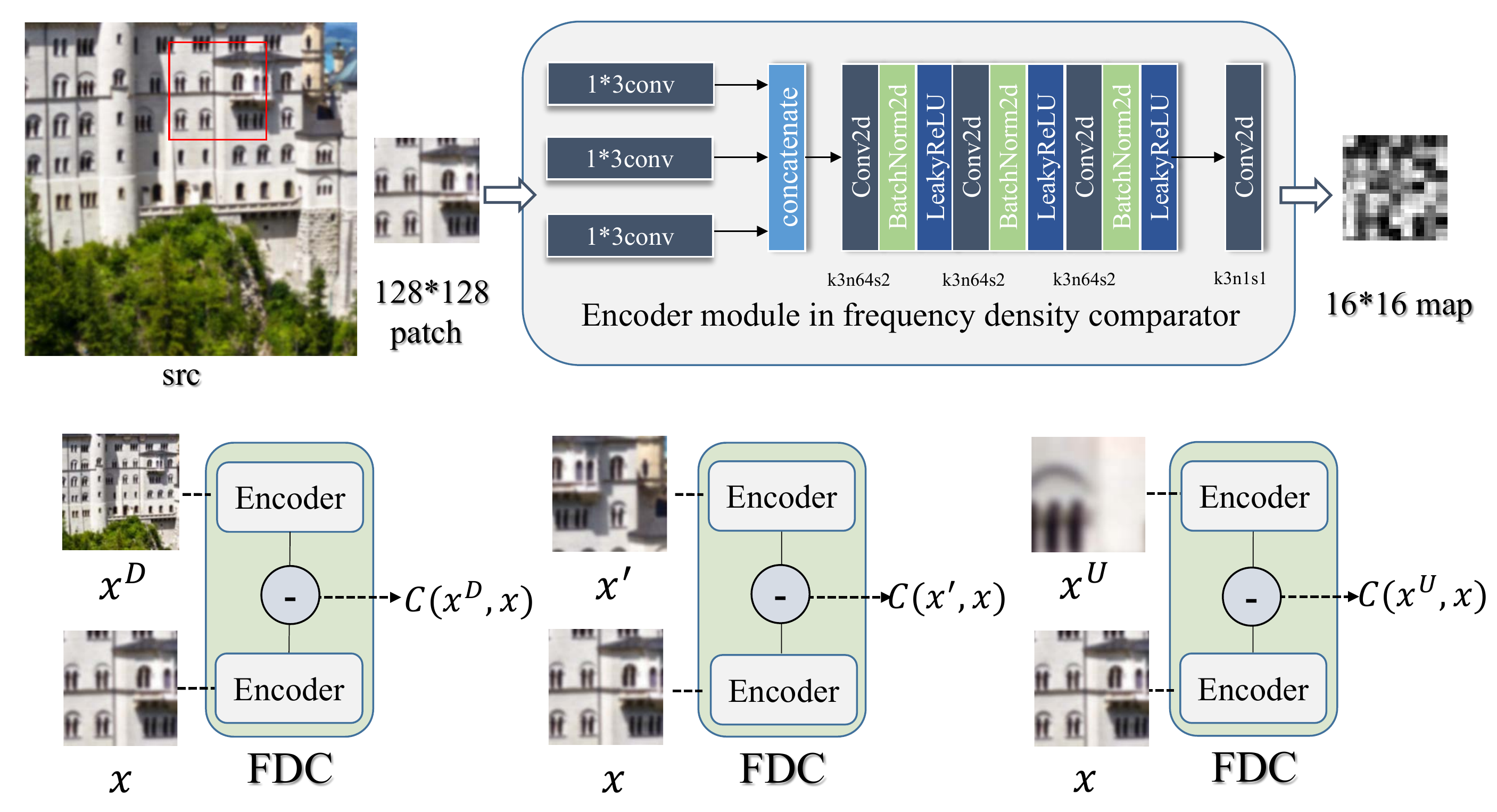} 
\caption{\textbf{Illustration of our frequency density comparator.} FDC has two encoder modules with shared weights. The encoder encodes the two input patches into latent space and outputs the embedding of frequency density. FDC then returns the substruction value of the two embeddings. 
}
\label{fig:FDC}
\end{figure}

As illustrated in Figure \ref{fig:FDC}, FDC is designed to capture the frequency density relationship of two input patches. For a real image ${\mathbf x}$, both down-sampling and up-sampling might change its frequency distribution as shown in Figure~\ref{fig:freq} (b). 
The density relations are as follows:
\begin{equation}
\begin{aligned}
C({\mathbf x}^D,~{\mathbf x})  &> 0\\
C({\mathbf x}^{'}~,~{\mathbf x})  &= 0\\
C({\mathbf x}^U,~{\mathbf x})  &< 0
\end{aligned}
\end{equation}
where $D$ and $U$ indicate down-sampling and up-sampling respectively. $C$ is the proposed comparator. ${\mathbf x}^{'}$ is another patch from source image.
The optimization of $C$ can be formulated as:
\begin{equation}
\begin{aligned}
& \arg \min _C \{~ \mathbb{E}_{{\mathbf x} \in S}[~\|C({\mathbf x}^D,{\mathbf x})-1\| \\
&+ \|C({\mathbf x}^{'},{\mathbf x})\| + \|C({\mathbf x}^U,{\mathbf x})+1\|~]~ \},
\end{aligned}
\end{equation}
where $S$ represents images from source domain.
At the beginning, FDC acquires the ability of comparing frequency density on coarse-grain according to the three kinds of patches.
To gradually obtain a fine-grained FDC, we dynamically narrow the classification boundary by adjusting the scales of up- and down-sampling during training. 

\paragraph{Frequency Consistency Loss}
FDC provides frequency consistency loss $\mathcal{L}_{FDC}$ for training the generator, which consists of three parts:
\begin{equation}
\begin{aligned}
\mathcal{L}_{FDC} &= \mathbb{E}_{{\mathbf x} \in S}[~\|C(G({\mathbf x}), {\mathbf x}^D)+1\| \\
&+ \|C(G({\mathbf x}),{\mathbf x})\| + \|C(G({\mathbf x}),{\mathbf x}^U)-1\|~].
\end{aligned}
\end{equation}
$\mathcal{L}_{FDC}$ ensures $G({\mathbf x})$ lies between the upper frequency boundary ${\mathbf x}^D$ and the lower boundary ${\mathbf x}^U$. Furthermore, the distance measurement between $G({\mathbf x})$ and ${\mathbf x}$ makes the kernel estimation closer to the real degradation.

\paragraph{Curriculum Learning Strategy}
In order to provide stable and accurate gradient information for the adaptation generator, we take the idea of curriculum learning ~\cite{bengio2013deep}. The training process of frequency density comparator is divided into different stages with increasing difficulty. 
The up- and down-sampling scale factors of source image for FDC training are set to multiple intermediate values dynamically approaching $1.0$.
We train FDC and adaptation generator simultaneously to ensure that the input patches of them share the similar frequency domain for each mini batch.

\subsection{Wavelet Discriminator}
In addition to $\mathcal{L}_{FDC}$, we also use adversarial loss to push LR closer to the source high-frequency domain.
Maintaining high-frequency information is very important for recovering image details. 
We adopt a similar idea with~\cite{fritsche2019frequency} and~\cite{wei2020unsupervised}, imposing adversarial loss only in the high-frequency space. 
High-frequency and low-frequency components are obtained by wavelet transform, but only the former is input into the discriminator.
Since only the non-semantic information of the image needs to be captured, the discriminator network has a shallow depth of only $4$ layers. Moreover, we use LSGAN~\cite{mao2017least} instead of original GAN. Denote the discriminator as $WD$, we optimize it as
\begin{equation}
\begin{aligned}
\arg \min_{WD} \{\mathbb{E}_{{\mathbf x} \in S}[(WD(G({\mathbf x})))^{2} + (WD({\mathbf x})-1)^{2}] \}.
\end{aligned}
\end{equation}
The adversarial loss fed back by the discriminator to the generator can be expressed as:
\begin{equation}
\mathcal{L}_{WD} =  \mathbb{E}_{{\mathbf x} \in S}[(WD(G({\mathbf x}))-1)^{2}],
\end{equation}
where $\mathcal{L}_{WD}$ denotes the wavelet discriminator loss. 

\subsection{Overall Loss}
As mentioned above, the overall loss $L_{total}$ contains two parts including frequency consistency loss $L_{FDC}$ and adversarial loss $L_{WD}$, which can be expressed as
\begin{equation}
\begin{aligned}
\mathcal{L}_{total} &= {\lambda}_{1} \cdot\mathcal{L}_{FDC} + {\lambda}_{2} \cdot \mathcal{L}_{WD},
\end{aligned}
\end{equation}
where ${\lambda}_{1}$/${\lambda}_{2}$ denotes weight of $\mathcal{L}_{FDC}$/$\mathcal{L}_{WD}$, respectively. 


\section{Experiments}

\newcommand{\tabincell}[2]{\begin{tabular}{@{}#1@{}}#2\end{tabular}}
\definecolor{background}{rgb}{0.9, 0.9, 0.9}
\begin{table*}[h]
    \centering
    \scriptsize
    \setlength{\tabcolsep}{2mm}{
\begin{tabular}{cccccc} \hline
        \hline
       \textbf{Type} & \textbf{Method} & \textbf{ISO.1} & \textbf{ISO.3} & \textbf{ISO.[1, 3] } &\textbf{ANI.}\\ \hline
       \multirow{3}*{\tabincell{c}{Ideal SR}} &ESRGAN finetuned &23.52~/~0.6333~/~0.5627 & 21.73~/~0.5773~/~0.6619& 22.48~/~0.5981~/~0.6142 &22.38~/~0.5960~/~0.6228\\
       & RCAN finetuned & 23.56~/~0.6349~/~0.5911 & 21.74~/~0.5789~/~0.6915 & 22.51~/~0.6011~/~0.6512 & 22.45~/~0.5991~/~0.6538 \\ 
       & ZSSR* & 23.54~/~0.6348~/~0.5811 & 21.73~/~0.5783~/~0.6779 & 22.49~/~0.6002~/~0.6400 & 22.40~/~0.5977~/~0.6444 \\
        \hline
       
        \multirow{3}*{\tabincell{c}{Correction SR}}& DeblurGANv2 w. RCAN & 23.69~/~0.6406~/~0.5791 & 21.74~/~0.5787~/~0.6895 & 22.74~/~0.6078~/~0.6351 & 22.73~/~0.6071~/~0.6357 \\ 
        & DeblurGANv2 w. ESRGAN & 23.61~/~0.6356~/~0.5538 & 21.98~/~0.5813~/~0.6399 & 22.65~/~0.6023~/~0.6057 &22.63~/~0.6011~/~0.6070 \\
       & Correction Filter w. RCAN & 23.06~/~0.6258~/~0.5791 & 21.59~/~0.5762~/~0.6769 & 22.23~/~0.5959~/~0.6386 & 22.19~/~0.5942~/~0.6406 \\ 
        \hline
       \multirow{2}*{\tabincell{c}{Blind SR}}

       & KernelGAN* &24.93~/~0.6787~/~0.4806 & 22.11~/~0.5847~/~0.6347 & 23.20~/~0.6192~/~0.5764 & 23.02~/~0.6133~/~0.5843 \\ 
       & IKC &\textcolor{blue}{26.74}~/~ \textcolor{blue}{0.7513}~/~0.3667  & \textcolor{blue}{22.15}~/~\textcolor{blue}{0.5890}~/~0.6700 & \textcolor{blue}{23.77}~/~\textcolor{blue}{0.6399}~/~0.5673 & \textcolor{blue}{23.57}~/~\textcolor{blue}{0.6368}~/~0.5754 \\ \hline
      
      \multirow{2}*{\tabincell{c}{FCA (ours)}}
      & \textbf{FCA w. ESRGAN*} & 24.30~/~0.6784~/~ \textcolor{red}{0.2533} & 22.01~/~0.5770~/~ \textcolor{red}{0.3795} &  20.47~/~0.5395~/~ \textcolor{red}{0.4008} & 20.80~/~0.5569~/~ \textcolor{red}{0.3739} \\ 
       & \textbf{FCA w. RCAN*} &  \textcolor{red}{27.35}~/~\textcolor{red}{0.7588}~/~\textcolor{blue}{0.3629} &
       
       \textcolor{red}{25.50}~/~ \textcolor{red}{0.6882}~/~\textcolor{blue}{0.4589} &
       
       \textcolor{red}{24.84}~/~ \textcolor{red}{0.6941}~/~\textcolor{blue}{0.4495} &
       
       \textcolor{red}{24.93}~/~ \textcolor{red}{0.6873}~/~\textcolor{blue}{0.4584} \\ \hline 
       \multirow{2}*{\textcolor{gray}{Upperbound}} & \textcolor{gray}{ESRGAN upperbound} &\textcolor{gray}{ 25.75~/~0.7034~/~0.1840} & \textcolor{gray}{22.62~/~0.5826~/~0.3071} & \textcolor{gray}{22.56~/~0.5853~/~0.2922}& \textcolor{gray}{22.70~/~0.5965~/~0.2960} \\ 
       & \textcolor{gray}{RCAN upperbound} & \textcolor{gray}{27.77~/~0.7738~/~0.3263} & \textcolor{gray}{25.62~/~0.6932~/~0.4415} & \textcolor{gray}{26.41~/~0.7236~/~0.3970} & \textcolor{gray}{26.23~/~0.7175~/~0.4072} \\ \hline
        \hline
    \end{tabular}}
    \caption{\textbf{Quantitative comparison results with $\mathbf{4\times}$ on various blur images.} The LR images are degraded with four different types of Gaussian kernels. The averaged [PSNR$\uparrow$ ~/~SSIM$\uparrow$ ~/~LPIPS$\downarrow$] measurement are reported. Unsupervised methods are marked with $*$. The best performance is shown in \textcolor{red}{red} and the second best in \textcolor{blue}{blue}.}
    \label{tab:synthetic}
\end{table*}

\subsection{Experiment Setup}
\paragraph{Training Setting}

We report training parameters setting here.
The network architecture is described as in Figure~\ref{fig:estimator} and Figure~\ref{fig:FDC}, where `k3n64s2' indicates that convolution kernel size, number of filters, and stride are set as $3, 64, 2$, respectively. The input size of adaptation generator is $512 \times 512$, and the scale factor is  $4\times$ which is the same as the SR factor. Gaussian kernels are of size $13 \times 13$ with maximum variance $9$. The down-/up-sampling scale factor during curriculum learning is decreasing from $3.5$ to $1.2$. 
In $\mathcal{L}_{total}$, we set $\lambda_{1}=1,\lambda_2=0.001$. 
We generate the HR images by
bicubically downscaling the source images with $2\times$, which can reduce blur effect~\cite{fritsche2019frequency}.

\paragraph{Datasets and Evaluation Metrics}
For synthetic experiments, we select the widely used DIV2K~\cite{timofte2017ntire} dataset, including $800$ training samples and $100$ validation samples as benchmark. 
For real-world experiment, we use the DPED~\cite{ignatov2017dslr} dataset containing $5,614$ training and $100$ testing images. 
Images in this dataset are more challenging due to its low-quality.
For synthetic images, we calculate PSNR, SSIM and LPIPS~\cite{zhang2018unreasonable} of different methods.
PSNR and SSIM focus on the fidelity of the image rather than visual quality, while LPIPS pays more attention to the similarity of the visual features.
For the case of real-world images, we mainly provide visual comparison due to no corresponding ground-truth images.

\subsection{Experiments on Various Blur Kernels}
In order to verify the effectiveness of our FCA, we use Gaussian blur kernel to generate non-ideal LR images. 
Note that the same degradation is performed on origin training images and down-sampled test images. 
We use four different types of degradation kernels with the same size $19 \times 19$:
\begin{itemize}
\item ISO.1: Isotropic Gaussian kernel with the fixed variance ($\sigma^{2}=1$) ;
\item ISO.3: Isotropic Gaussian kernel with the fixed variance ($\sigma^{2}=3$) ;
\item ISO.[1, 3]: Isotropic Gaussian kernel with unfixed variances ($\sigma^{2} \in [1,\ 3]$);
\item ANI.: Anisotropic Gaussian kernel ($\sigma_x^{2},\sigma_y^{2} \in [1,\ 3],\theta \in [0,\ 2\pi]$).
\end{itemize}


\begin{figure*}[!ht]
\centering
\includegraphics[width=0.9\textwidth]{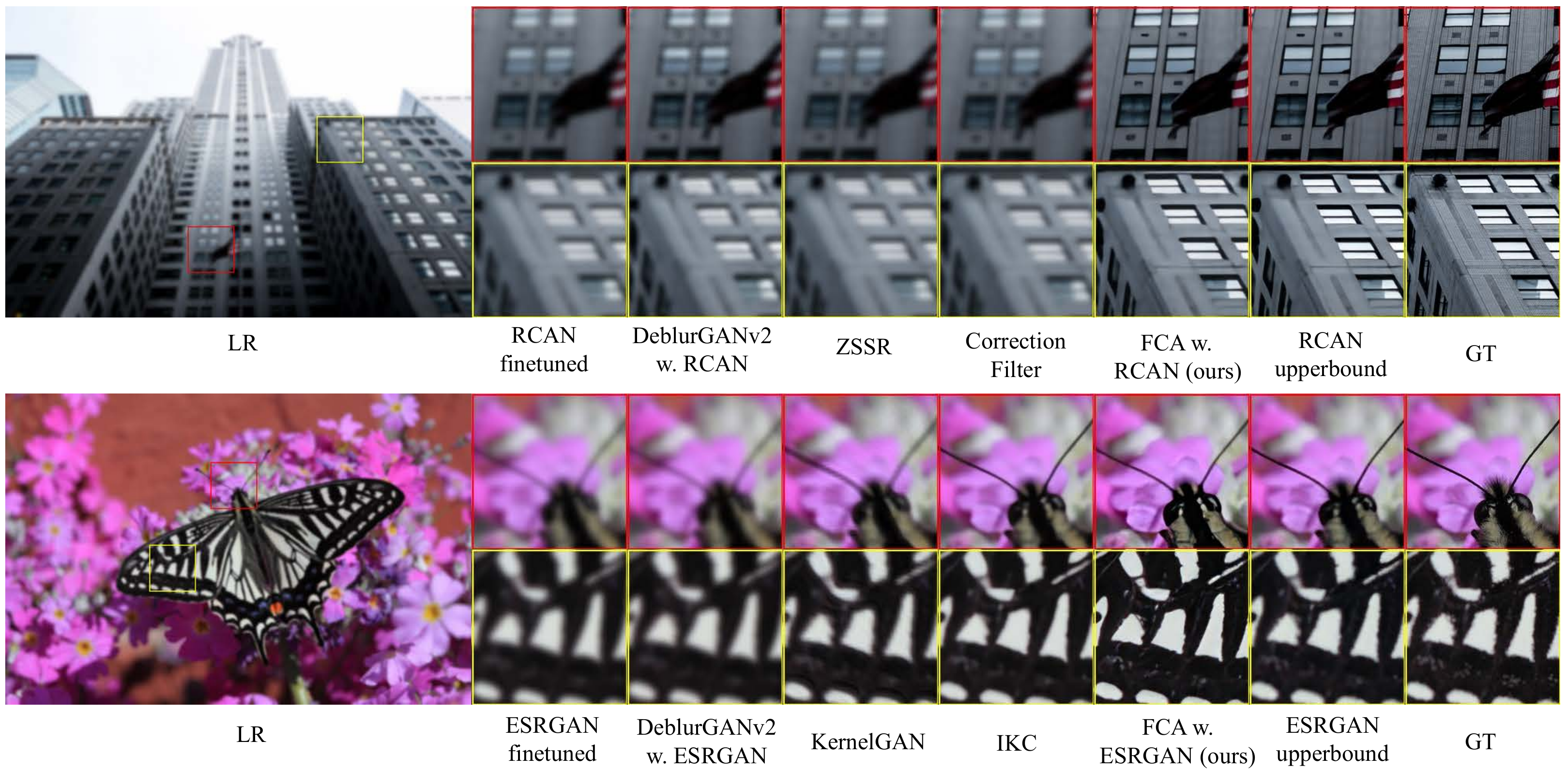}
\caption{\textbf{Visual results with $\mathbf{4\times}$ compared with state-of-the-art SR methods on blur images with ISO.1.}}
\label{fig:synthetic}
\end{figure*}

We train FCA on four types of dataset synthesized with kernels above, then construct HR-LR pairs for RCAN~\cite{zhang2018image} and ESRGAN~\cite{wang2018esrgan}. Our method is noted as `FCA w.'.
For comparison, we also show the SR performance upper bound noted as `Upperbound'. 
Under this setting, training pairs are undegraded HRs and LRs degraded using the ground-truth kernels. 
Since FCA is to obtain domain-consistent training LRs, `Upperbound' is the performance upper bound it can reach. 

\paragraph{Comparison with Ideal SR}
In this comparison, we verify that the proposed FCA improves the performance of ideal SR model (\textit{i.e.}, RCAN and ESRGAN).
We finetune the pretrained model on the pairs constructed according to their original way (i.e. bicubic) from source datasets, and note this method as `finetuned'. 
Experimental results in Table~\ref{tab:synthetic} show that `FCA w. RCAN' achieves the best PSNR/SSIM performance and `FCA w. ESRGAN' achieves the best LPIPS on all the four types of degradation kernels. The proposed FCA improves RCAN/ESRGAN with a large margin indicating its consistent adaptation into source domain is effective. 

\paragraph{Comparison with Correction SR}
In this part, we compare our FCA with deblurring method (\textit{i.e.}, DeblurGANv2~\cite{kupyn2019deblurgan}) and domain correction method (\textit{i.e.}, Correction Filter~\cite{hussein2020correction}). Different from estimating internal degradation of image, these methods try to remove degradation from LR image and restore it to ideal state. We then classify these methods as `Correction SR'. As described in~\cite{zhang2019deep}, deblurring first works better than doing SR first. We then combine DeblurGANv2 with RCAN and ESRGAN, Correction Filter with RCAN respectively for comparison. 
Quantitative results in Table~\ref{tab:synthetic} show that these methods fail to reduce the domain gap.
It is harder to recover the high-frequency components from the LR than degrading a clean one. Visual results in Figure~\ref{fig:synthetic} also confirms that FCA succeeds to recover realistic details with higher perceptual quality by accurate frequency adaptation.  
\paragraph{Comparison with Blind SR}
Among blind SR methods, KernelGAN~\cite{bell2019blind} and IKC~\cite{gu2019blind} are representative methods. KernelGAN estimates the specific degradation kernel for ZSSR~\cite{shocher2018zero} to generate SR results.
For comparison, we run ZSSR with the bicubic kernel and put it under the type `Ideal SR'.
IKC predicts the kernel and generates SR results in an iterative way.
Quantitative and qualitative results are displayed in Table~\ref{tab:synthetic} and Figure~\ref{fig:synthetic}. 
KernelGAN promotes the PSNR performance about $1.0$ compared with ZSSR. However, its degradation kernel estimation is not accurate, thus the SR process still suffers from frequency domain gap. Benefiting from correct guidance of the proposed FDC, our method achieves much better quality metric results than these representative blind methods.
From the visual results, we can see that important low-frequency structures and realistic high-frequency details are reconstructed successfully. 

\begin{figure*}[!ht]
\centering
\includegraphics[width=0.8\textwidth]{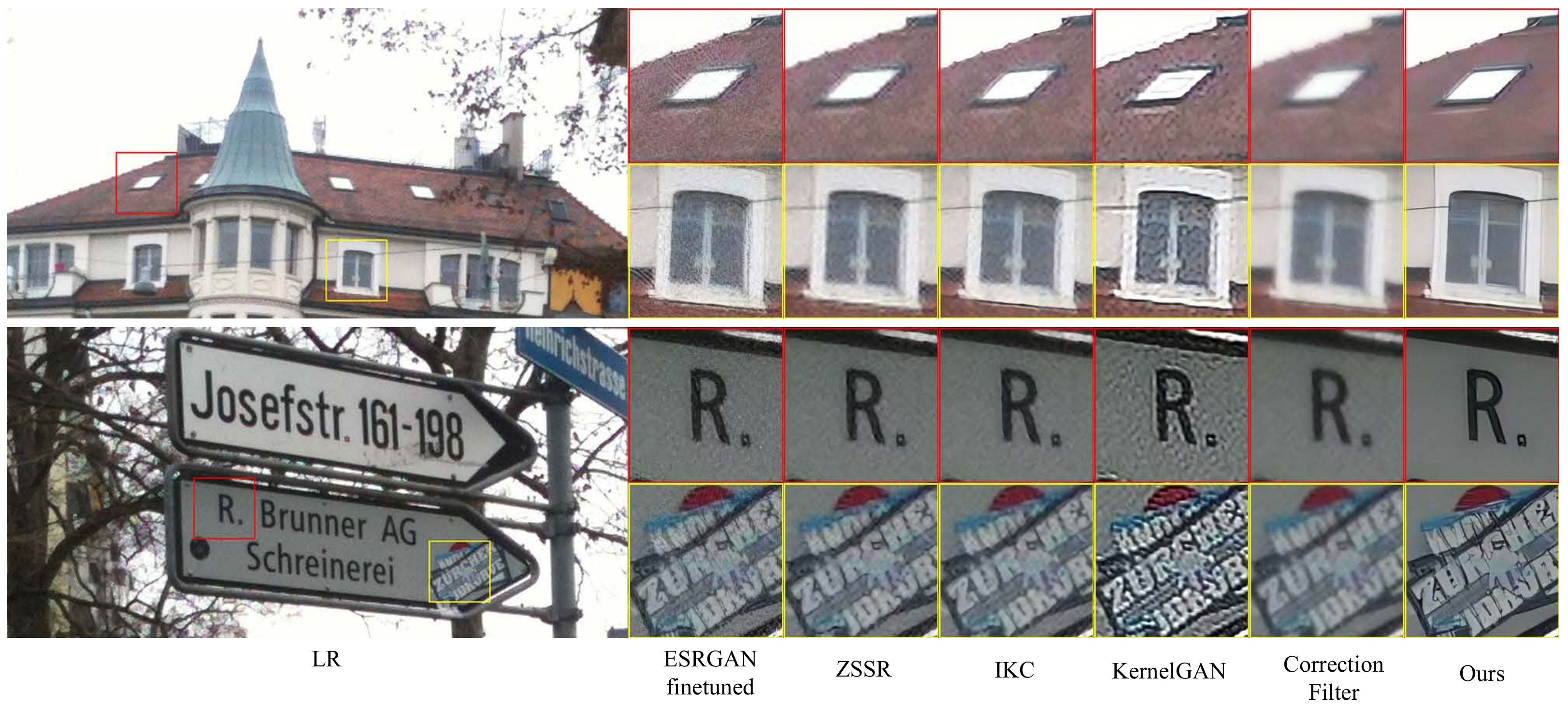} 
\caption{\textbf{Visual results with $\mathbf{4\times}$ compared with state-of-the-art SR methods on real images.}} 
\label{fig:real}
\end{figure*}

\subsection{Ablation Study}
\begin{table}[h!]
    \centering
    \tiny
    \setlength{\tabcolsep}{2mm}{
  \begin{tabular}{cccc} \hline
        \hline
      Kernels & \textbf{$\mathcal{L}_{WD}$} & \textbf{$\mathcal{L}_{FDC}$} & \textbf{$\mathcal{L}_{WD}\ \&\  \mathcal{L}_{FDC}$} \\ \hline
      ISO.1&27.07~/~0.7543~/~0.3688 & \textbf{27.59}~/~\textbf{0.7680}~/~\textbf{0.3448}& 27.35~/~0.7588~/~0.3629\\
      ISO.3& 23.23~/~0.6275~/~0.5361  &25.45~/~0.6861~/~0.4647 & \textbf{25.50}~/~\textbf{0.6882}~/~\textbf{0.4589}\\ 
      ISO.[1,3]& 23.98~/~0.6520~/~0.5199  &24.63~/~0.6888~/~0.4446 & \textbf{24.84}~/~\textbf{0.6940}~/~\textbf{0.4500}\\ 
      ANI.& 24.44~/~0.6590~/~0.5100 & 24.82~/~0.6850~/~0.4630 & \textbf{24.93}~/~\textbf{0.6870}~/~\textbf{0.4580} \\ \hline
        \hline
    \end{tabular}}
    \caption{\textbf{Ablation study of the proposed frequency consistency loss and the adversarial loss.} [PSNR/SSIM/LPIPS] results of three different models trained with different loss functions setting on $4\times$ synthetic images are listed. The best performances are marked in \textbf{bold}.} 
    \label{table:Ablation}
\end{table}
Furthermore, in order to analyze the effects of
the proposed frequency consistency loss and the adversarial loss
on the performance of the estimator, we conduct ablation experiments under the condition that only one loss is introduced into the generator on synthetic LR images. The results in Table~\ref{table:Ablation} show that FDC plays a key role in the domain adaptation. Meanwhile, the final performance is improved from good cooperation with wavelet discriminator.


\subsection{Experiments on Real World Images}

\begin{figure}[t!]
\centering
\includegraphics[width=0.9\columnwidth]{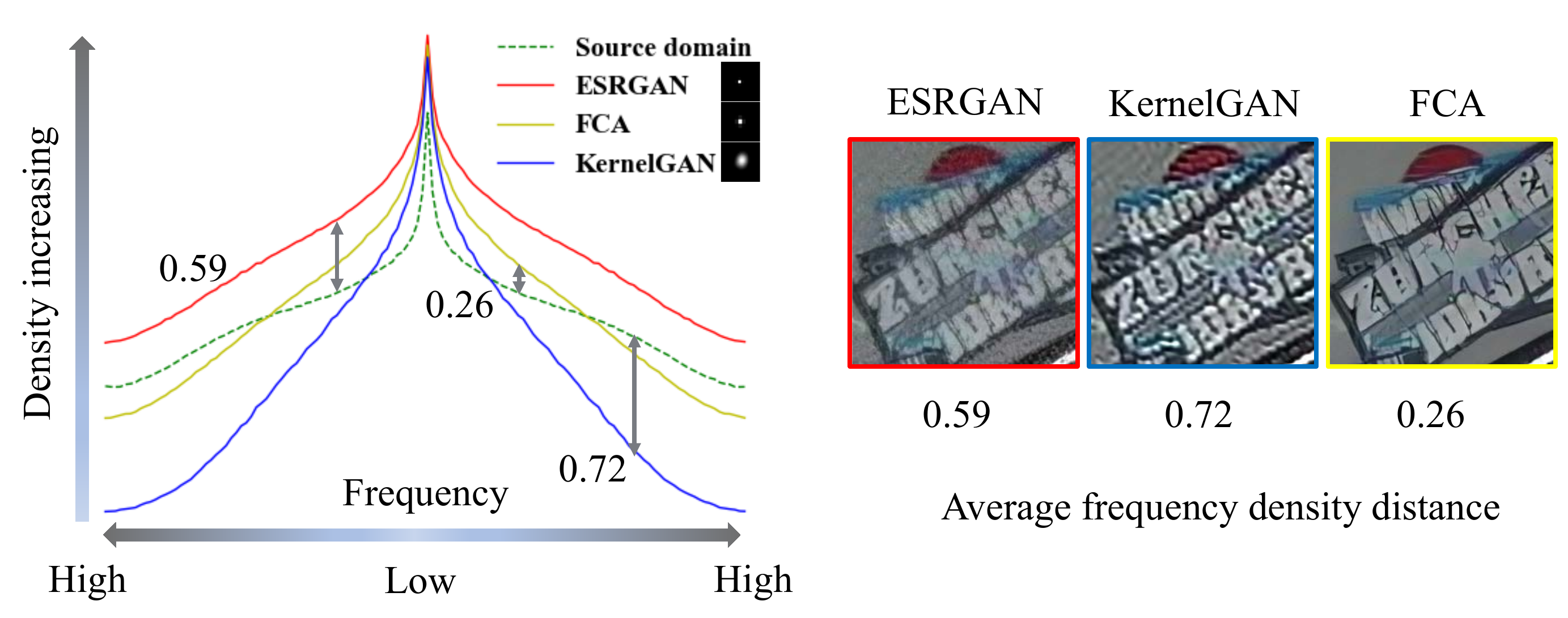} 
\caption{\textbf{Frequency density distributions of degraded LR and SR results by different methods on real-world images.} The frequency distribution distance indicates that ESRGAN and KernelGAN generate inconsistent degraded images. Our FCA estimates proper kernels, thus obtaining LRs whose domain is closer to source domain.}
\label{fig:real_freq}
\end{figure}

In the experiment on real-world dataset, we combine FCA with ESRGAN because its perception-oriented loss function helps to reconstruct richer details. 
For fair comparison, we finetune ESRGAN for the same iterations. 
Other comparative methods include ZSSR, KernelGAN, IKC, and Correction Filter. 
From the visual comparison in Figure~\ref{fig:real}, we notice that ESRGAN finetuned, ZSSR, IKC, Correction Filter generate over-smoothing results, while KernelGAN generates over-sharpening results. These blurry results that lacks of high-frequency details fail to enhance important edges (\textit{e.g.}, windows and letters). In contrast, our results show clear dividing line between two different surfaces. On the other hand, KernelGAN produces many undesirable artifacts though it looks good at first, which suggests its inaccurate estimation of degradation. We also provide no-reference assessment comparison results with the winning method (Impressionism~\cite{ji2020real}) in NTIRE 2020 Challenge on Real-World Image Super-Resolution~\cite{lugmayr2020ntire}. Our result is $14.5$ on PIQE~\cite{venkatanath2015blind}, $17.6\%$ lower than $17.6$ of Impressionism.

To provide quantitative measurement of frequency density distance, we propose to calculate it as follows: 
\begin{equation}
\begin{aligned}
\bar{\mathcal{D}}(X,Y) &= \frac{1}{n} \Sigma_l [ |\mathcal{F}_X(l) - \mathcal{F}_Y(l) | ],
\end{aligned}
\end{equation}
where $\bar{\mathcal{D}}(X,Y)$ denotes the distance between domain $X$ and $Y$, $n$ is the number of frequency.
By FCA's accurate estimation, the degraded LRs are close to source domain as shown in Figure~\ref{fig:real_freq}, thus obtaining more natural result. 


\section{Conclusion}
In this paper, we propose a novel frequency consistent adaptation for real-world super-resolution, which keeps the degraded images and the original ones consistent in the frequency domain. In the proposed FCA, we carefully design a frequency domain density comparator to estimate the degradation of the source domain through an unsupervised training method. Experiments on synthetic and real-world datasets show that the proposed FCA is effective in real-world SR, avoiding performance dropped caused by incorrect degradation. As a general unsupervised degradation estimation method, FCA can be combined with easy-implemented SR models and achieves state-of-the-art performance on real images. Experiments show our SR results achieve higher fidelity and better visual perception.

\section{Acknowledgments}
This work is supported by the Natural Science Foundation of China under Grant 61672273 and Grant 61832008.

\bibliography{main}
\end{document}